\documentclass{preprint}
\usepackage{graphicx} 
\usepackage[colorlinks=true]{hyperref}
\usepackage{booktabs}
\usepackage[hypcap=false]{caption}
\usepackage{makecell}
\usepackage{multirow}
\usepackage{tabularx}
\usepackage{amsmath, amsthm, amssymb, amsfonts}
\usepackage{thmtools}
\usepackage{setspace}
\usepackage{geometry}
\usepackage{float}
\usepackage{hyperref}
\usepackage[utf8]{inputenc}
\usepackage[english]{babel}
\usepackage{framed}
\usepackage[dvipsnames]{xcolor}
\usepackage{tcolorbox}
\usepackage{algpseudocode}
\usepackage{algorithm}

\usepackage[numbers,comma,sort&compress]{natbib}

\title{A MapReduce Approach to Effectively Utilize Long Context Information in Retrieval Augmented Language Models}

\author[1]{Gongbo Zhang}
\author[2]{Zihan Xu}
\author[3]{Qiao Jin}
\author[1]{Fangyi Chen}
\author[1]{Yilu Fang}
\author[4]{Yi Liu}
\author[5,6]{Justin F. Rousseau}
\author[7]{Ziyang Xu}
\author[3]{Zhiyong Lu}
\author[1,*]{Chunhua Weng}
\author[2,*]{Yifan Peng}

\affil[1]{Department of Biomedical Informatics, Columbia University, New York, NY, USA}
\affil[2]{Department of Population Health Sciences, Weill Cornell Medicine, New York, NY, USA}
\affil[3]{National Center for Biotechnology Information, National Library of Medicine, National Institutes of Health, Bethesda, MD, USA}
\affil[4]{Division of Endocrinology, Diabetes and Metabolism, Department of Medicine, Weill Cornell Medical College, New York, NY, USA}
\affil[5]{ Department of Neurology, University of Texas Southwestern Medical Center, Dallas, TX, USA}
\affil[6]{Peter O'Donnell Jr. Brain Institute, University of Texas Southwestern Medical Center, Dallas, TX, USA}
\affil[7]{Department of Dermatology, NYU Grossman School of Medicine, New York, NY, USA}
\affil[*]{Corresponding author(s). Email(s): \url{cw2384@cumc.columbia.edu}, \url{yip4002@med.cornell.edu}}

\begin{document}

\maketitle

\begin{abstract}
While holding great promise for improving and facilitating healthcare, large language models (LLMs) struggle to produce up-to-date responses on evolving topics due to outdated knowledge or hallucination. Retrieval-augmented generation (RAG) is a pivotal innovation that improves the accuracy and relevance of LLM responses by integrating LLMs with a search engine and external sources of knowledge. However, the quality of RAG responses can be largely impacted by the rank and density of key information in the retrieval results, such as the ``lost-in-the-middle'' problem. In this work, we aim to improve the robustness and reliability of the RAG workflow in the medical domain. Specifically, we propose a map-reduce strategy, BriefContext, to combat the ``lost-in-the-middle'' issue without modifying the model weights. We demonstrated the advantage of the workflow with various LLM backbones and on multiple QA datasets. This method promises to improve the safety and reliability of LLMs deployed in healthcare domains.
\end{abstract}

\section{Introduction}\label{introduction}

Large language models (LLMs) are finding their way into an expanding range of healthcare domains, holding tremendous potential for improving patient care, enhancing communication and education, and facilitating clinical workflow effectiveness~\cite{Singhal2023-jc, Jin2023-ka, Haupt2023-df, Peng2023-zh, Zhang2024-oi, Idnay2024-as, Jin2024-gq}. LLMs are useful for answering common queries related to diseases or personal risk, interpreting laboratory results, and getting advice on medical condition management \cite{Spotnitz2024-fm, Zelin2024-er, Tian2023-at, Zhang2024-eq, Park2024-db}. Despite the potential of LLMs, the deployment of LLMs in healthcare faces serious safety concerns. LLMs struggle to generate accurate and up-to-date responses on current topics, due to outdated knowledge, lack of domain-specific expertise, or hallucination \cite{Cao2020-js, Chen2024-hw, Raunak2021-zi, Ji2023-rp, Zhang2024-bu}.

Retrieval-Augmented Generation (RAG) is a pivotal innovation to enhance the quality and relevance of responses in large language models (LLMs) \cite{Guu2020-gy, Lewis2020-ll, Borgeaud2022-mx, Izacard2020-oz}. Typically, a RAG system consists of a retrieval module and a generative module. When a user query is provided as input, the system first uses the retrieval module to fetch relevant documents or data snippets by searching through external data sources. Next, the generative module takes the retrieved information as input and produces a response to the user query. With the help of the retrieval module, the generative module can provide more accurate and factual answers without the need for continual training or fine-tuning. As such, RAG poses a promising direction for applications requiring high factual accuracy and specificity \cite{Chen2024-hw, Ding2024-ab}.

However, prompting LLMs with contextual information has trade-offs. On the one hand, providing contextual information enhances the model's ability to perform the downstream tasks by augmenting LLMs with external domain-specific knowledge that is under-represented in their pretraining data. On the other hand, the input of LLMs is bounded by the limit of their context windows. Even though recently released models can process an increasing number of tokens, the increased amount of content to reason over can still hinder model performance \cite{Li2024-xy}. The quality of RAG completion also depends on the retrieval results, such as the density or positions of query-relevant information \cite{Chen2024-hw, Ding2024-ab, Liu2023-hz, Jiang2023-dq, Xiong2024-ng}. As retrieval systems are still imperfect, it is inevitable to retrieve information irrelevant to the user query \cite{Chen2024-hw}.

A recent study reports an issue of ``lost-in-the-middle'', i.e., the position of key information in the LLM context impacts the quality of the model completions \cite{Liu2023-hz}. This issue occurs when a lengthy context of information is retrieved, and the highly relevant information is not ranked at the top or bottom of the retrieval results. We refer to these positions as spotlight positions, and the document containing key information as the key document. If not ranked at the top, the key information may be neglected by the generative module, resulting in incomplete or inaccurate responses to the user queries \cite{Liu2023-hz}. How to effectively utilize contextual information in RAG applications remains to be an open research question. Current studies attribute this issue to positional attention bias, i.e., more attention weights are allocated more to information at spotlight positions than others \cite{He2023-xj, Hsieh2024-jb}. To address the issue, existing methods mainly focus on adjusting the model weights, either by fine-tuning LLMs \cite{He2023-xj} or directly adjusting the attention weights \cite{Hsieh2024-jb}. However, adjusting model weights can lead to catastrophic forgetting \cite{Kemker2018-lc, Kirkpatrick2017-lu}, i.e., the overall performance of LLMs degrades upon adopting new information on a specific task.

In this research, we aim to address the ``lost-in-the-middle'' issue without modifying model weights. Our strategy involves increasing the density of key information within the context, rather than modifying model weights. The lower bound for RAG is closed-book settings, where LLMs have access only to the question with no extra information. The upper bound is Oracle settings, where only relevant key information is provided in the context. Compared to closed-book settings, LLMs perform significantly better in Oracle settings. These two scenarios represent opposite ends of the spectrum concerning key information density. In closed-book settings, the density is essentially zero because no external information is provided. In contrast, Oracle settings boast nearly 100\% density, as only relevant information is supplied. RAG sits in the middle, where relevant information is often mixed with irrelevant content. We hypothesize that the density of key information affects downstream model performance.

Therefore, we propose a novel framework, BriefContext, to transform the long-context reasoning task into multiple short-context reasoning tasks. The core of the framework leverages the map-reduce concept \cite{Dean2010-gr, Zhang2023-gj, Bergui2024-cf, Senthamil-Selvi2024-ao, Mv2024-hb, Luo2024-qf}, originally designed for processing massive data in parallel. In our workflow, we divide the long context into multiple partitions and dispatch them to multiple LLM sessions. The additional LLM service requests incur extra costs. However, suppose the key document is returned at the top of the ranking. In that case, the extra cost is unnecessary since the downstream generative module can already take advantage of the key information at spotlight positions. To avoid unnecessary costs, we introduce a preflight mechanism to predict the occurrence of ``lost-in-the-middle''. Such a task is challenging since the key document is unknown beforehand. Here, we employ a heuristic based on consistency across different ranking results to predict the occurrence of the issue.

We evaluated the proposed framework via both controlled experiments and integration testing. In particular, we evaluate general-purpose LLMs on answering medical QA questions that require domain knowledge in depth \cite{Singhal2023-jc, Xiong2024-ng, Lievin2024-uu, Jin2021-mn, Tsatsaronis2015-xd, Jin2019-hd}. This choice of models and dataset exemplifies the scenarios where knowledge encoded from pretraining data is insufficient to answer the questions well. Our controlled experiments changed the position of key information and compared BriefContext with a regular RAG pipeline. In the integration testing, we use the ranking order from the real-world retrieval results. These experiments demonstrate that BriefContext consistently outperforms the RAG baseline by a substantial margin when the key information is placed in the middle. BriefContext also improves the model performance when the key information is placed in spotlight positions.

Furthermore, to understand how BriefContext improves the RAG pipeline, we investigate the following questions, each of which corresponds to a module in the pipeline:

(1) Can LLMs resolve conflicts correctly when the LLM context contains conflicting information? We find LLMs can correctly resolve 74.7\% of cases with conflicting information in the context window. Consequently, BriefContext achieved a higher overall accuracy than the vanilla RAG.

(2) Do LLMs utilize short context more effectively than long context? Here, we prove the hypothesis that with the same key information in the context, LLMs perform better at reasoning over shorter contexts than longer ones. We controlled the number of documents in the context information and evaluated LLMs in different settings. We find that model performance decreases as the number of documents increases, even though the same key information is present in the context. This confirms that short context is utilized more effectively than long context in RAG. Furthermore, since LLMs perform better at reasoning over shorter contexts than longer ones, the problem of reasoning over long context can be divided into multiple subtasks of reasoning over short context, and the correct answer can be more easily located in one of the subtasks.

(3) How well does the preflight check predict the occurrence of ``lost-in-the-middle''? We show that the preflight check can predict the issue occurrence with a recall of 92.61\% but a precision of 50.18\%. About 35.7\% of true-negative cases can be correctly filtered by the preflight check.

(4) What is the relationship between the retrieval results and the positional attention bias? We show that positional attention bias is triggered when the key documents contain similar vocabulary to other documents in the context that do not provide supporting information to the user query.

\section{Results}\label{results}

\subsection{BriefContext Overview}\label{briefcontext-overview}

Our goal is to mitigate the issue of ``lost-in-the-middle'', which affects the performance of RAG in QA tasks. This issue arises when the sequence of document retrieval influences the quality of the information extracted and used in generating responses.

\begin{figure}
    \centering
    \includegraphics[width=0.7\linewidth]{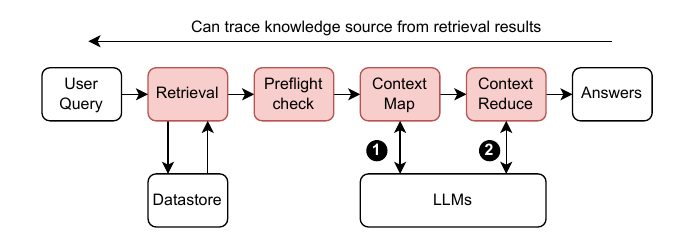}
    \caption{Workflow of BriefContext. In the Context Map operation (1), the retrieved documents are divided into multiple partitions to create multiple RAG subtasks. In the Context Reduce operation (2), the responses were collected from the previous step and summarized into a final response.}
    \label{fig:Workflow}
\end{figure}

Our proposed BriefContext consists of four modules (Figure \ref{fig:Workflow}): Retrieval, Preflight check, Context Map, and Context Reduce. The Model Development section provides a detailed description. The workflow initiates when a user inputs a query. This query is converted into an encoded representation and used to search a knowledge base, where documents have been previously encoded into vectors using the same encoder (Retrieval module). Then, the retrieved documents are sorted using two distinct algorithms: MedCPT and BM25. It is important to note that MedCPT is also used in the primary retrieval module. By comparing the two rankings, we develop the Preflight check module to conjecture the existence of the ``lost-in-the-middle'' issue. If the issue is detected, the ContextMap module engages. Here, the retrieved documents are partitioned. Using partitions created in the ContextMap step, the LLMs are prompted to extract relevant information from each partition. Furthermore, the extracted responses are collected and injected into the ContextReduce module. Here, the aggregated responses undergo a summarization process to distill the most pertinent information. Finally, the summarized information is formatted into a coherent response and provided to the user as the final answer.

This workflow is designed to minimize the detrimental effects of retrieval order by reshaping how information is processed and integrated from various sources. By doing so, the BriefContext enhances the accuracy and reliability of responses in QA tasks, ensuring that users receive precise and relevant information regardless of how the initial data was retrieved.

We tested the workflow on multiple-choice questions, which allow scalable evaluation. The multiple-choice questions are all publicly available. Specifically, we chose the MIRAGE benchmark for this purpose. For a comprehensive test, we also evaluated the workflow on open-ended questions generated using publicly available education materials. The details are described in the Method section.

\subsection{Can we address the issue of ``lost-in-the-middle'' without changing model weights?}\label{can-we-address-the-issue-of-lost-in-the-middle-without-changing-model-weights}

\begin{figure}
    \centering
    \includegraphics[width=\linewidth]{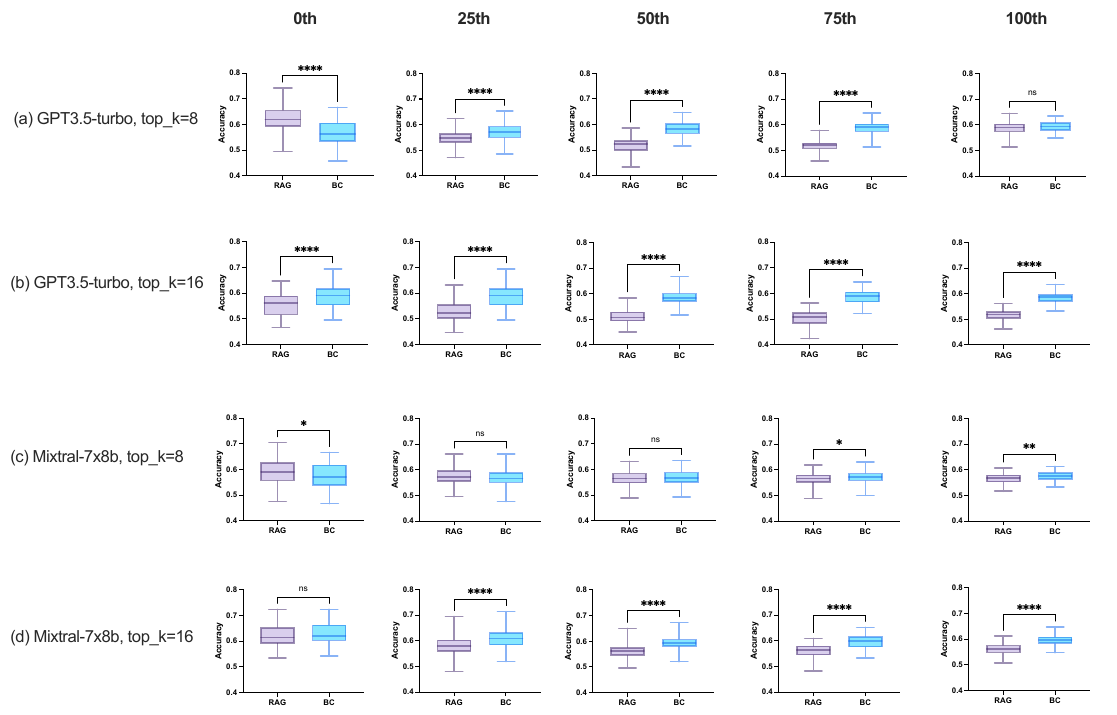}
    \caption{Relationship between QA accuracy and positions of key information in the LLM context: (a-b) GPT-3.5-Turbo, (c-d) Mixtral-7x8b. The quartiles refer to the positions where the key document is located. Significance levels: * - $p < 0.05$; ** - $p < 0.01$; *** - $p < 0.001$; **** - $p < 0.0001$; ns - Not significant.}
    \label{fig:Relationship}
\end{figure}

To answer this question, we evaluated BriefContext in both controlled studies with synthetic rankings and integration testing with real-world rankings. In the controlled study, we used all of the PubMed articles and a collection of textbooks \cite{Tsatsaronis2015-xd} that are widely used by medical students as the knowledgebase. While a portion of the knowledge base or corpus where the dataset was derived (e.g., PubMed abstracts or textbooks) is probably included in the pre-training of LLMs, we deem the comparison remains fair, since we used the same backbone LLMs for RAG and BriefContext. We selected 20\% of questions from PubMedQA \cite{Jin2019-hd}, and MedCPT \cite{Jin2023-ka} as the primary search engine. The evaluation metric is accuracy, which is the ratio of correctly answered questions. As shown in Figure \ref{fig:Relationship} and Supplementary Table \ref{sup tab:Relationship}, BriefContext utilizes the external information in the middle of the context more effectively than the baseline RAG workflow. Using Mixtral-7x8b \cite{Jiang2024-jw} as the LLM backbone, the accuracy averaged over different positions was improved from 56.76 to 57.53 when the top 8 documents were included in the prompt ($top\_k = 8$) and from 56.38 to 60.00 when $top\_k = 16$. Using GPT-3.5-turbo \cite{Brown2020-le} as the LLM backbone, the accuracy was improved from 53.33 to 59.51 when $top\_k =8$, and 52.0 to 58.86 when $top\_k = 16$.

\begin{figure}
    \centering
    \includegraphics[width=0.5\linewidth]{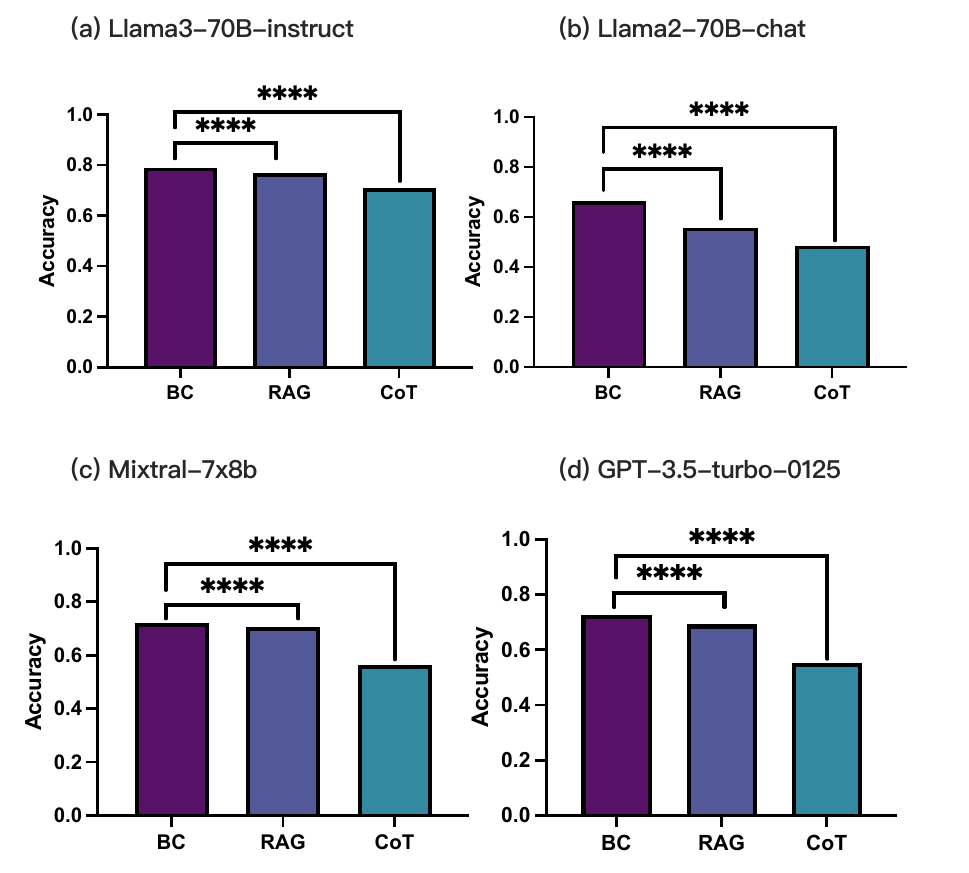}
    \caption{Integration testing of BriefContext with different LLM backbones: (a) Llama3-70B-instruct, (b) Llama2-70B-chat, (c) Mixtral-7x8b, and (d) GPT-3.5-turbo-0125. BC - BriefContext. RAG - Retrieval-augmented generation. CoT - Chain-of-Thought. Significance levels: * - $p < 0.05$; ** - $p < 0.01$; *** - $p < 0.001$; **** - $p < 0.0001$; ns - Not significant.}
    \label{fig:Integration}
\end{figure}

We used the baseline Chain-of-Thought (CoT) and RAG accuracies reported in the MIRAGE benchmark. The results of integration testing shown in Figure \ref{fig:Integration} and Supplementary Table \ref{sup tab:Number} demonstrate that BriefContext has improved the overall accuracy across different LLM backbones. With LLama2-70B-chat, the accuracy was improved from 55.81 to 66.47; with LLama3-70B-instruct, the accuracy was improved from 76.75 to 79.03; with Mixtral-7x8b, the accuracy was improved from 70.52 to 72.20; with GPT-3.5-turbo-0125, the accuracy was improved from 69.19 to 72.51. We also invited three medical experts to evaluate model responses to 48 open-ended medical questions. Out of the 48 questions, our method generates better answers than the RAG baseline for 29.2\% of questions and worse answers for 12.5\% of questions. For the remaining 58.3\% of questions, our method and the RAG baseline produced the same responses.

\subsection{Can LLMs resolve the conflicts in the retrieved external knowledge in the ContextReduce step?}\label{can-llms-resolve-the-conflicts-in-the-retrieved-external-knowledge-in-the-contextreduce-step}

In the BriefContext workflow, we divided the long text into multiple partitions. One issue is that LLM answers based on different context partitions are not always the same. We refer to such a situation as context with conflict information. It's unclear how LLMs deal with such a context. To investigate this problem, we used 20\% of PubMedQA questions with synthesized rankings. The experimental setup, including the knowledge base, search engine, and backbone LLMs, is the same as the above control studies of BriefContext.

\begin{figure}
    \centering
    \includegraphics[width=.7\linewidth]{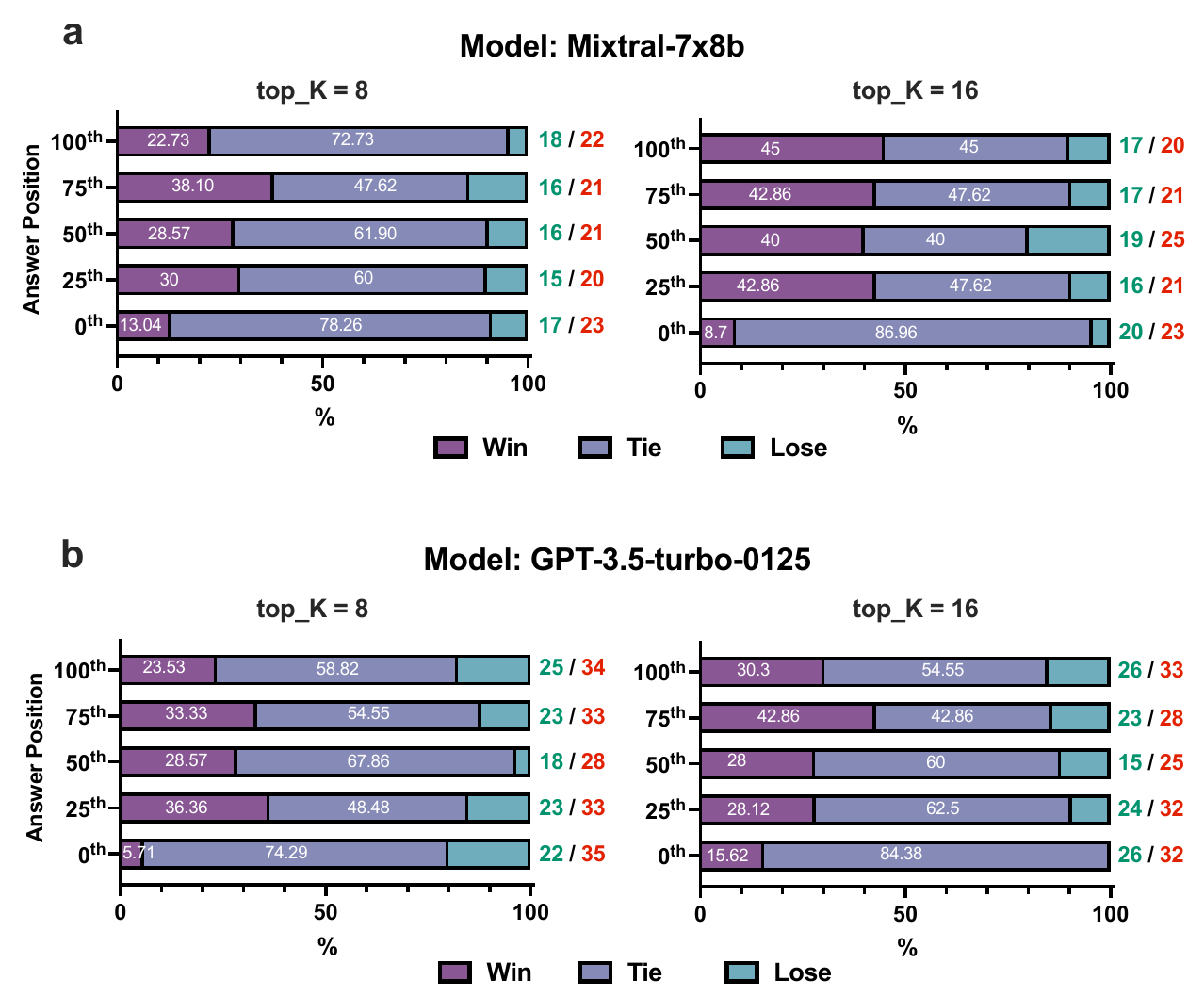}
    \caption{Number of cases (red) with conflict information provided to LLMs and number of correctly resolved cases (green).}
    \label{fig:Number}
\end{figure}

The results are shown in Figure \ref{fig:Number}. Overall, Mixtral-7x8b resolved 171 out of 217 cases with conflicting contextual information correctly; GPT-3.5-turbo resolved 225 out of 313 cases. We also reported the win/tie/lose ratio (defined in the Method section) details in Supplementary Table \ref{sup tab:Number}. Overall, BriefContext consistently demonstrates a higher win rate than lose rate, which indicates the advantages of BriefContext handling context with conflict information. The advantages are more manifested in 25\textsuperscript{th}, 50\textsuperscript{th}, 75\textsuperscript{th} percentile of positions than the others. This also highlights that the key information is under-utilized by the vanilla RAG, especially when the context contains conflicting information.

\subsection{Do LLMs favor short context over long context in the ContextMap step?}\label{do-llms-favor-short-context-over-long-context-in-the-contextmap-step}

\begin{figure}
    \centering
    \includegraphics[width=.7\linewidth]{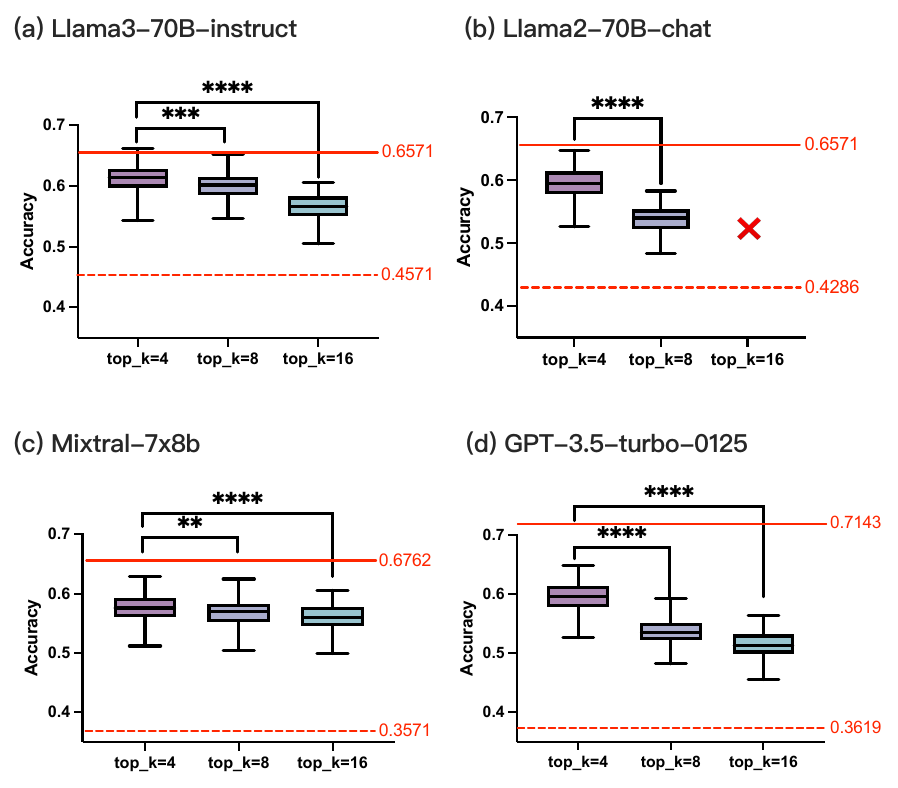}
    \caption{Medical QA accuracy of LLMs with various numbers of documents as context information. The top solid line shows the performance in the Oracle settings. The bottom dotted line shows the performance of CoT. With the same key document in the context, the accuracy decreases as the number of documents increases. (a) Llama3-70B-instruct, (b) Llama2-70B-chat, (c) Mixtral-7x8b, and (d) GPT-3.5-turbo-0125. BC - BriefContext. RAG - Retrieval-augmented generation. CoT - Chain-of-Thought. Significance levels: * - $p < 0.05$; ** - $p < 0.01$; *** - $p < 0.001$; **** - $p < 0.0001$; ns - Not significant.}
    \label{fig:Medical}
\end{figure}

To answer this, we used the same questions, knowledge base, and search engine as in the question above with synthetic rankings. We strategically placed key documents at different positions in the context (i.e., retrieved PubMed abstracts) and reported the average accuracy. We evaluated the same 4 LLM backbones using various numbers ($top\_k$) of documents in the context. Figure \ref{fig:Medical} shows that the LLMs favor short over long context.

\subsection{Can the occurrence of ``lost-in-the-middle'' be predicted by the Preflight check?}\label{can-the-occurrence-of-lost-in-the-middle-be-predicted-by-the-preflight-check}

It typically remains unknown which documents contain the key information. It's thus unclear whether the ``lost-in-the-middle'' issue happens or not. To predict the occurrence of the issue, we used the consistency across different ranking results as a heuristic (see details in the Methods section). We evaluate how well the heuristic can predict the issue. We define consistency as the IoU rate between rankings from MedCPT and BM25. The threshold is set to 0.2. When the IoU is larger than 0.2, we posit that MedCPT has placed the key document at top positions, i.e., the ranking issue does not occur. To validate this hypothesis, we used precision, recall, and F1, where a true-positive is defined as an issue of ``lost-in-the-middle'' that happened and was successfully captured using the IoU score. The test was performed using queries from PubMedQA and BioASQ datasets, and results from PubMed were retrieved using MedCPT. The IoU heuristic achieved 50.18\% precision, 92.61\% recall, and 65.09\% F1 (Supplementary Table \ref{sup tab:Confusion}).

\subsection{What is the relationship between positional attention bias and retrieval results?}\label{what-is-the-relationship-between-positional-attention-bias-and-retrieval-results}

Recent studies \cite{He2023-xj, Hsieh2024-jb} pointed out that lost-in-middle-issue is attributed to the positional attention bias, i.e., models exhibit U-shaped attention patterns where documents at the beginning or end of the inputs receive higher attention values, regardless of their relevance. We argue that positional attention bias is related to inaccurate retrieval results that are irrelevant to the user query but contain vocabulary similar to the key documents. Recall that most modern LLM architectures employ self-attention, which calculates pair-wise inner product of embeddings as attention weights \cite{Vaswani2017-jd}. Each position is typically represented as a concatenation of position and text embedding vectors \cite{Devlin2018-ct, Radford2018-xs}. We hypothesize that positional attention bias is triggered only when the text embeddings of key documents are similar to other documents in the context. In other words, the positional attention bias will disappear when the key document can be associated with the query successfully and distinguished clearly from other retrieved documents in the context.

\begin{figure}
    \centering
    \includegraphics[width=0.7\linewidth]{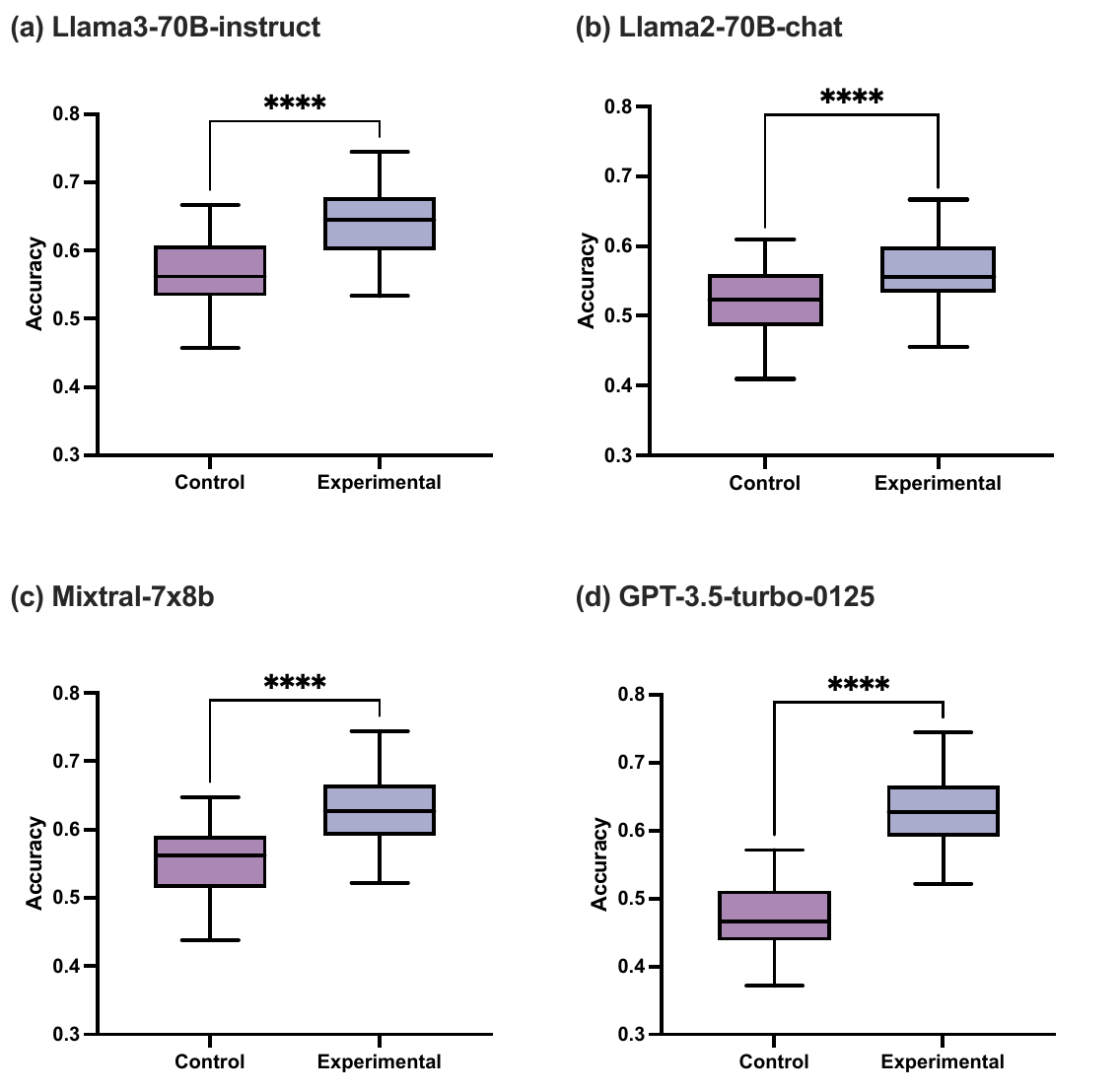}
    \caption{Relationship between QA accuracy and different context setups. In the Control group, all documents come from results returned by MedCPT. In the experimental group, the context consists of key documents and others selected at random from the knowledge base. (a) Llama3-70B-instruct, (b) Llama2-70B-chat, (c) Mixtral-7x8b, and (d) GPT-3.5-turbo-0125ontext. RAG - Retrieval-augmented generation. CoT - Chain-of-Thought. Significance levels: * - $p < 0.05$; ** - $p < 0.01$; *** - $p < 0.001$; **** - $p < 0.0001$; ns - Not significant.}
    \label{fig:RelationshipQA}
\end{figure}

To prove this hypothesis, we randomly selected 20\% of multiple-choice questions (n=105) from the PubMedQA dataset. We set up two search engines to retrieve documents relevant to the questions. In the control group, we used MedCPT as the search engine and retrieved the top 16 documents from the external knowledge base using the input query. In the experimental group, we synthesize retrieval results by mixing the key documents with documents randomly selected at random from the knowledge base. The randomly selected documents were highly likely irrelevant to the input query. To manifest the lost-in-the-middle issue, we place the key document right in the middle of the LLM context for both groups. We provide the two retrieval results to downstream LLMs as contextual information and report the accuracy. Figure \ref{fig:RelationshipQA} shows that the accuracy is higher when the key documents are mixed with random documents (experimental group) as compared to relevant documents (control group), even though the key documents are placed right in the middle of the context. These results prove that positional attention bias is overpowered by text-embedding-based attention when the key information is distinguishable from other documents in the context. Furthermore, this observation highlights a limitation of search engines based on embedding representation or dense retrieval. These search engines sometimes return irrelevant documents that manifest a high resemblance to the query vocabulary.

\section{Discussion}\label{discussion}

Our experiments demonstrate that BriefContext improves the robustness regarding the order of retrieved documents in the RAG paradigm without adjusting model weights. Our proposed workflow improved accuracy on several biomedical QA datasets. This is demonstrated via both controlled studies and integration testing, as shown in Figures 2 and 3. When conflicting information is present in the context, Mixtral-7x8b correctly resolved 78.8\% of the cases with conflicting information in the context, while GPT-3.5-turbo resolved 71.8\% of the cases, as shown in Figure \ref{fig:Number}. As such, the BriefContext can better utilize the key document than RAG, mainly when the context contains conflicting information. However, LLMs do not always correctly resolve the conflict information. Here, we illustrate one example where BriefContext fails, but vanilla RAG succeeds. Consider the question with ID 18507507 in PubMedQA, ``\emph{The promise of specialty pharmaceuticals: are they worth the price?}''. The publication record (PMID 18507507), labeled as the key information, supports a positive answer. Other retrieved records present irrelevant information, which results in an answer with a lower level of certainty (e.g., PMID 28911475, PMID 24991326). Such retrieved records can lead to a positive answer in one partition and an uncertain answer in another. In the phase of ContextReduce, the backbone LLM favored the uncertain answer, leading to errors. Despite this, in most cases the conflicting information can be resolved correctly.

In addition to the LLM capabilities of correctly resolving most conflicting information, we also show that key information can be better utilized in a short context than in a long context. To prove this, we construct various sets of context information with varying numbers of documents but the same key information. As shown in Figure \ref{do-llms-favor-short-context-over-long-context-in-the-contextmap-step}, the QA accuracy decreases as the number of documents is increased. By dividing a long list of documents into multiple batches, we decompose a challenging RAG task into multiple subtasks with shorter context. Resolving the ``lost-in-the-middle'' issue is also attributed to this division operation, which is defined as the ContextMap operation in our pipeline. In cases where the key documents are ranked at the spotlight positions, the vanilla RAG workflow can already utilize the key information. However, it is challenging to predict where the key document is ranked without knowing which document contains the key information. To combat this issue, we propose a preflight check mechanism to predict the ``lost-in-the-middle'' occurrence. Supplementary Table \ref{sup tab:Confusion} shows that the preflight check achieves 50.18\% precision, 92.61\% recall, and 65.09\% F1.

Earlier studies pointed out that the issue of ``lost-in-the-middle'' is attributed to positional attention bias \cite{He2023-xj, Hsieh2024-jb}. In this study, we show that positional attention bias only manifests when the key documents are not distinguishable from other documents in the context based on topic similarity to the query. The positional attention bias can be overpowered by the segment embeddings when the key documents are distinguishable. As shown in Figure \ref{fig:RelationshipQA} (experimental group), the key documents can be effectively utilized even if placed right in the middle of the context. This highlights the limitations of embedding-based search engines, which mainly rely on superficial lexical similarity to perform the retrieval task without deeply understanding the relationship between user queries and the returned documents \cite{Steck2024-eh}.

We identify the following sources of medical QA errors in the RAG paradigm. First, LLMs sometimes resolve conflicting information incorrectly. Although about 78.8\% and 71.8\% of conflicting information were resolved by Mixtral-7x8b and GPT-3.5, respectively, they failed to provide correct answers for the rest of the cases, resulting in wrong final answers. Second, although the RAG paradigm improves LLMs via external knowledge sources, we show that LLMs may still fail to answer questions correctly even in the oracle settings, where only key documents were provided as the context. While this issue is beyond the scope of the ``lost-in-the-middle'', this highlights the gap between the RAG paradigm and the strict requirement for accuracy in the medical domain.

Our experiment has a few limitations. Firstly, due to the lack of open-ended questions annotated with key documents, we cannot quantitatively evaluate the impact of key document positioning on QA responses. However, we addressed this by conducting a controlled experiment using multiple-choice questions where the key document was strategically placed at various positions within the prompt context. Secondly, our choice of the off-the-shelf LLMs without any modifications presents another limitation. A future direction of this work could explore the context map-reduce paradigm with fine-tuned or task-specific LLMs. Lastly, our current focus is on QA tasks in the medical domain. In future studies, we plan to explore the application of LLMs to other tasks and QA tasks in other scientific domains.

\section{Conclusion}\label{conclusion}

We propose BriefContext, a map-reduce approach, to effectively utilize long context in RAG workflow for answering questions in the medical domain. First, we showed that LLMs can better utilize short context than long context. Next, by dividing the long context into several subtasks, we improve the model performance on biomedical QA tasks without adjusting model weights. To avoid unnecessary extra costs on LLMs service, we then introduced a preflight check mechanism to prognose the ranking issue without knowing which document contains key information. We show LLMs can correctly resolve 74.7\% of cases with conflicting information in the context window. BriefContext takes advantage of this capability of LLMs and shorter context, which explains how BriefContext improves biomedical QA accuracy in RAG workflow. Lastly, we discussed when positional attention bias is triggered. We hope this assists future research on the root cause of the positional attention bias.

While our proposed BriefContext framework was evaluated only within the biomedical question-answering in this study, it shows promise for generalizing to tasks that require effective processing of long-context data, such as extracting pertinent data from lengthy, duplicative electronic health records, legal document analysis, historical research, or technical report summarization. Future studies could explore these applications to evaluate the generalizability and adaptability of BriefContext in addressing diverse and complex information retrieval challenges.

\section{METHODS}\label{methods}

We describe the methods in detail in four main sections, aligning with the study aims and the Results section.

\subsection{Data}\label{data}

\paragraph{Multiple-choice questions.} To develop the model and ensure its scalable evaluation, we used multiple-choice questions, where the correctness of model outputs can be determined without necessitating further expert feedback. We chose the MIRAGE \cite{Xiong2024-ng} benchmark for this purpose, which consists of three medical examination QA subsets (MMLU-Med \cite{Hendrycks2020-oi}, MedQA-US \cite{Jin2021-mn}, and MedMCQA \cite{Pal2022-rm}) and two biomedical research QA subsets (PubMedQA \cite{Jin2019-hd} and BioASQ-Y/N \cite{Tsatsaronis2015-xd}) (Supplementary Table \ref{sup tab:dataset}).

Given that our goal is to improve RAG pipelines, we specifically used two biomedical subsets (PubMedQA and BioASQ-Y/N), due to their reliance on external knowledge databases that can augment the capabilities of LLMs. Furthermore, to maintain a diversity of question types, we used MedMCQA, the largest medical examination QA dataset \cite{Pal2022-rm}.

\paragraph{Open-ended questions.} In the real-world practice of medical QA, questions always arise without predefined options, reflecting the open-ended nature of real-world scenarios. As such, we present MedQ, a dataset comprising 48 open-ended questions. We created these questions using StatPearls \cite{StatPearls2024-av}, a source that summarizes up-to-date medical knowledge and practice across various specialties. In particular, we selected articles focusing on neurology, endocrinology, and dermatology.

To formulate the questions, we prompt GPT-4 to generate pairs of PICO (participant, intervention/comparison, and outcomes) questions and answers. The generated QA pairs were then reviewed by three specialties (dermatology, neurology, and endocrinology) to ensure their accuracy and relevance.

\paragraph{External knowledge base.} Following the practice of this benchmark work \cite{Xiong2024-ng}, we built a knowledge base with components: (1) The entire collection of abstracts indexed in PubMed, and (2) a set of 18 medical textbooks \cite{Jin2021-mn} that are widely used by medical students and serve as preparation materials for the USMLE exams (Supplementary Table \ref{sup tab:Components})\footnote{\url{https://github.com/jind11/MedQA}}.

\subsection{Model Development}\label{model-development}

\paragraph{Retrieval.} Given a large collection of documents $\mathbb{D}$, the main goal of the retrieval module is to select a subset of documents \(D_{r} = \{ d_{1},\ d_{2},\ ...\ d_{k}\} \subset \mathbb{D}\) relevant to the user query \emph{Q}, where \(k\) is the number of retrieved documents. To perform an effective and efficient retrieval, we first encode each document \(d_{i}\) and the query \(Q\) into numerical vectors of the same fixed dimension, denoted as \(embed(d_{i})\) and \(embed(Q)\), respectively.

\paragraph{Preflight Check.} The collection is then sorted by the relevance to the query. We denote the resultant ranking as \(R_{LLM} = [r_{1}^{LLM},r_{2}^{LLM},...,r_{k}^{LLM}]\). where the relevance of a document \(d_{i}\) to a query is determined by the inner product of the two embedding vectors \(r_{i}^{LLM} = embed(d_{i})^{\top}embed(Q)\). Based on the ranking results, we discuss two possible outcomes. First, when the key information is ranked at the top positions, the generative module can take advantage of the retrieved information. In this case, there is no need to include too many articles in the context. Several results ranked at the top provide enough information to answer the question. Second, when the key information is ranked beyond the spotlight positions, the key information is probably to be neglected.

To combat ranking related issues, a common approach is to employ hybrid rankings, which ensemble several ranking results into a new order using reciprocal ranking fusion (RRF). While RRF demonstrated advantages in end-to-end RAG evaluation, there is no guarantee that documents with key information will always be placed at top positions in the hybrid ranking results. This new ranking still leaves the ``lost-in-the-middle'' issue unresolved. It's also unrealistic to expect any retrieval system to always place the documents of interest at the very first position.

Inspired by the hybrid ranking algorithms, we use the consistency among different retrieval systems to conject the occurrence of ranking issues without knowing which documents contain key information. In particular, we calculate the intersection-over-union (IoU) rate between the top \(n\) results. In addition to the retrieval system based on dense representation of documents, we use another ranking algorithm, BM25, to rerank the documents in \(R_{LLM}\). The new ranking is denoted as \(R_{BM25} = [r_{1}^{BM},r_{2}^{BM},...,r_{k}^{BM}]\). Next, we conduct a preflight check to determine whether to invoke the BriefContext subroutine in the RAG pipeline. The preflight check is formally defined as an indicator function,

\begin{equation}
(R_{LLM},R_{BM25},n)=
\begin{cases}
1, & \text{if } \frac{R_{LLM}[:n]\cap R_{BM}[:n]}{R_{LLM}[:n]\cup R_{BM}[:n]}\\
0, & \text{otherwise}
\end{cases}
\end{equation}

The choice of the threshold is crucial in balancing the trade-off between precision and recall. While it's possible to further optimize the precision and overall F1 by adjusting the threshold, we chose a value that ensures high recall. This decision is based on the fact that false positive errors result in extra cost, while false negative errors could leave the errors in vanilla RAG unaddressed. To better demonstrate the effectiveness of our methods, we prioritize achieving high recall over high precision.

\paragraph{ContextMap.} The ContextMap operation divides \(D_{r}\) into a partition \(P(D_{r})\) (i.e., the sets in \(P\) are subsets of \(D_{r}\), and the elements of \(P\) are mutually exclusive) and converts each subset as a prompt. Here, each subset has the same number of documents, denoted as \(D_{r}^{s} \in P\). The output is a list of prompts with the same instruction and user query, as outlined in Supplementary Algorithm \ref{alg:map}. Consider a partition of \(D_{r} = \{ d_{1},\ d_{2},\ ...\ d_{8}\}\) as \(D_{r\ }^{1} = \{ d_{1},\ d_{2},\ d_{3},d_{4}\}\) and \(D_{r}^{2} = \{ d_{5},\ d_{6,},d_{7},d_{8}\}\), the resultant prompts are ``$\{instruction\} \{query\} [doc 1] d_{1} [doc 2] d_{2} \ldots$'' and ``$\{instruction\} \{query\} [doc 1] d_{5} [doc 2] d_{6} \ldots$''. The only difference between the prompts is the contextualized documents. It has been pointed out that decoder-only models cannot attend to query tokens if the query is only placed behind the contextual information, since decoder-only models only attend to prior tokens by each timestamp. To combat this effect, we adopt query-aware contextualization, where a prompt consists of instruction, context information, and the user query placed before the context. Since not all documents in \(D_{r}\) are necessarily related to the query \(Q\), we instruct the model to either extract the relevant information or truthfully report no detection of any relevant information. The operation of ContextMap can be processed in parallel via multi-threading, where each thread formats a prompt. This batch processing is straightforward to implement since the prompt formatting subroutine only requires read access to the context.

\paragraph{ContextReduce.} After the context mapping, we next query the backbone LLM to extract relevant information from the context and answer the user query, as outlined in Supplementary Algorithm \ref{alg:reduce}. The relevant information is autoregressively sampled from the probability distribution over the model vocabulary conditioned on the instruction, query, and provided context:

\begin{equation}
y_{t}^{info} \sim p_{\theta}(Q, D_{rs}^{S}, I_{e}, y_{0:t - 1}^{info})
\end{equation}

where we denote the model weights as \(\theta\), extraction instruction \(I_{e}\), query \(Q\), shard of context \({D_{rs}}^{S}\), and \({y_{t}}^{info}\) the sampled information. The invocations to extraction can also be streamlined in parallel. The extracted information is then used to generate a summarization prompt, where we provide instructions \(I_{s}\) to ignore empty information. The final answer is also directly sampled from the probability distribution over the model vocabulary conditioned on the summarization instruction, extracted information, and query:

\begin{equation}
y_{t}^{answer} \sim p_{\theta}(Q, y^{info}, I_{s}, y_{0:t - 1}^{answer})
\end{equation}

As in a typical map-reduce workflow, the long context of relevant documents is first divided and dispatched to worker LLMs to create requests for extracting relevant information. After all the worker LLMs finish their processing jobs, they return to the LLM allocator to aggregate the individual results.

Below, we discuss the extra cost incurred by invoking the BriefContext subroutine. Here, we use the pricing model of most proprietary LLMs, e.g., GPTs, where users are charged by the number of input and output tokens. We denote the number of tokens in the prompt instruction and context as \emph{N\textsubscript{ins}} and \emph{N\textsubscript{con}}, and the maximum number of output tokens as \emph{N\textsubscript{out}}, respectively. The prices of input and output per token are denoted as \emph{p\textsubscript{input}} and \emph{p\textsubscript{output}}. The context is divided into M partitions. The cost of vanilla RAG is

\begin{equation}
O(N_{con} \cdot p_{input}\  + N_{ins} \cdot p_{input} + \ N_{out} \cdot \ p_{out})
\end{equation}

while the cost of BriefContext invocation is

\begin{equation}
O(N_{con} \cdot p_{input}\  + M \cdot N_{ins} \cdot p_{input} + \ (M + 1) \cdot N_{out} \cdot p_{out})
\end{equation}

Since the lengths of instruction and output are much shorter than the context information, the extra cost incurred by BriefContext invocations is not significant in scale.

In our cost analysis, we adopted the big-O notation, proving that BriefContext and vanilla RAG are at the same level in terms of theoretical complexity. However, in real-world scenarios, constant factors do play a role. For example, the extra prompt can account for up to 10\% of input tokens. While these extra tokens do not impact the big-O analysis, they do result in about a 10\% increase in actual costs. It's challenging to accurately quantify the percentage increase in cost, since this varies by the specific prompt, retrieved documents, and batch size in BriefContext.

Another factor to consider is the occurrence of ``lost-in-the-middle'' issues, which can vary by the queries, corpus, and choice of retrieval models. To help understand the frequency of these issues, we reported the average number of tokens per request, with 8 publication records retrieved for each query. In BriefContext, the average numbers of input and output tokens per request are 5,496.5 and 247.5, respectively. In vanilla RAG, these numbers are 3066.0 and 183.0. Given these numbers, we hypothesize the preflight check can help reduce extra costs by up to 35\%.

\subsection{Evaluation}\label{evaluation}

\textbf{Can we address the issue of ``lost-in-the-middle'' without changing model weights?} To answer this question, we evaluated BriefContext in both controlled studies with synthetic rankings and integration testing with real-world rankings. In the controlled study, we used the same experimental setup as the above question, i.e., the knowledge base of PubMed articles and textbooks, the 20\% subset of questions from PubMedQA, and MedCPT as the primary search engine. The evaluation metric is accuracy. We synthesized rankings by placing key information at different positions, including 0\textsuperscript{th}, 25\textsuperscript{th}, 50\textsuperscript{th}, 75\textsuperscript{th}, and 100\textsuperscript{th} percentile of positions in the context. We used Mixtral-7x8B and GPT-3.5-turbo as LLM backbones since these two models benefit more from retrieval augmentation than others (Figure \ref{fig:Relationship}). We compared the BriefContext with the vanilla RAG workflow using the same backbone LLM and external knowledge as the context in the prompts.

In the first integration testing, we used all questions from MedMCQA \cite{Pal2022-rm}, PubMedQA \cite{Jin2019-hd}, and BioASQ-Y/N \cite{Tsatsaronis2015-xd} from the MIRAGE \cite{Xiong2024-ng} benchmark dataset. The evaluation metric is accuracy. We selected LLama2-70B-chat  \cite{Touvron2023-we}, LLama3-70B-instruct \cite{meta2024-ss}, Mixtral-7x8b \cite{Jiang2024-jw}, and GPT-3.5-turbo \cite{Brown2020-le} as backbone LLMs, all of which have been used in the published benchmark results \cite{Xiong2024-ng}. We used the baseline closed-book (CoT) and RAG accuracies that were reported in the MIRAGE benchmark results \cite{Xiong2024-ng}. We used the same knowledge base as in the controlled studies. The knowledge base is a subset of the corpus that was used in the MIRAGE benchmark results reported by Xiong et al. \cite{Xiong2024-ng}. Our knowledge base thus contains no extra information as compared to theirs, which makes a fair comparison between BriefContext and RAG. In BriefContext, we used MedCPT as the search engine. The order of retrieved documents by MedCPT was kept the same when the prompt context was constructed. The top\_k is set to 16. In the second integration testing, we invited three medical experts to help curate 48 open-ended question-answer pairs from their specialty domain and compare our method with the RAG baseline.

\textbf{Can LLMs resolve the conflicts in the retrieved external knowledge in the ContextReduce step?} To investigate this problem, we used 20\% of PubMedQA questions with synthesized rankings. The experimental setup, including the knowledge base, search engine, and the backbone LLM, is the same as the above experiments. We define the occurrence of conflict information as an event in which LLMs return inconsistent answers given different context partitions of the same query results. We further define that the conflict is correctly resolved if the final answer is correct. We report the number of cases with conflict information and how many cases were correctly resolved by our proposed workflow. We also compare BriefContext with the vanilla RAG, which has the same backbone LLM in these cases. The comparison results consist of three possible outcomes: 1) our method wins the comparison if it resolves the conflict information correctly while the RAG baseline answers the question incorrectly; 2) the lose outcome is defined similarly; or 3) the outcome is a tie when both BriefContext and RAG answer the question either correctly or incorrectly.

\textbf{Do LLMs favor short context over long context in the ContextMap step?} To answer this, we used the same questions, knowledge base, and search engine as in the question above with synthetic rankings. We strategically placed key documents at the 0\textsuperscript{th}, 25\textsuperscript{th}, 50\textsuperscript{th}, 75\textsuperscript{th}, and 100\textsuperscript{th} percentile of the positions in the context (i.e., retrieved PubMed abstracts) and calculated the accuracy averaged over the five positions.

\textbf{Can the occurrence of ``lost-in-the-middle'' be predicted by the Preflight check?} To predict the occurrence of the issue, we used the consistency across different ranking results as a heuristic. We evaluate how well the heuristic can predict the issue. In this experiment, we selected all questions from PubMedQA and BioASQ, where the answers were also annotated with the PMID of articles that contained the key information. The issue occurrence is defined as an event where the key document is ranked beyond the top N positions. The threshold N is set to 3 since model performance drops significantly when N becomes larger than 3, according to earlier studies \cite{Liu2023-hz, He2023-xj, Hsieh2024-jb}. We used the same knowledge base as in the above questions. We used MedCPT as the primary search engine and BM25 as the secondary search engine to rerank the retrieval results from MedCPT. We define consistency as the IoU rate between rankings from MedCPT and BM25. The threshold is set to 0.2.

\textbf{What is the relationship between positional attention bias and retrieval results?} In our study, we decouple the impact of segment embeddings on attention weights from the impact of positional embeddings. Recall that Transformer architecture adopts the self-attention mechanism, where the weight is calculated as an inner-product between each pair of embeddings \cite{Vaswani2017-jd}. Each embedding consists of positional, token, and segment embeddings, which encode position and semantics, respectively \cite{Borgeaud2022-mx, Devlin2018-ct}. We randomly selected 20\% of multiple-choice questions (n=105) from the PubMedQA dataset. Each question is a multiple-choice question, and the evaluation metric is accuracy. The knowledge base consists of two components. One is all of the abstracts indexed at PubMed\footnote{\url{https://pubmed.ncbi.nlm.nih.gov/}}, and the other is a collection of 18 textbooks \cite{Jin2021-mn} that medical students widely use for preparing USMLE. We used MedCPT \cite{Jin2023-ka} as the search engine to obtain relevant information from the knowledge base. MedCPT was specifically pretrained on biomedical literature using user click information \cite{Jin2023-ka}.

\paragraph{Data availability:}

The data and codes underlying this article will be available upon request.


\paragraph{Funding/Support:} This project was sponsored by the National Library of Medicine grant R01LM009886, R01LM014344, and the National Center for Advancing Clinical and Translational Science awards UL1TR001873 and UL1TR002384. Q.J. and Z.L are supported by the NIH Intramural Research Program, National Library of Medicine. We also want to express our gratitude to Amazon Web Services (AWS) for providing the computational resources used in our research.

\paragraph{Role of the Funder/Sponsor:} The funder had no role in the design and conduct of the study; collection, management, analysis, and interpretation of the data; preparation, review, or approval of the manuscript; and decision to submit the manuscript for publication.















\paragraph{Author Contributions:} Study concepts/study design, \textbf{G.Z., C.W., Y.P.}; manuscript drafting or manuscript revision for important intellectual content, all authors; approval of final version of the submitted manuscript, all authors; agrees to ensure any questions related to the work are appropriately resolved, all authors; literature research, \textbf{G.Z., Y.P.}; experimental studies, \textbf{G.Z., Z.X., Q.J., F.C., Y.F.}; human evaluation, \textbf{Y.L., J.F.R., Z.X.}; data interpretation and statistical analysis, \textbf{G.Z., Y.P.}; and manuscript editing, all authors.


\paragraph{Conflict of Interest Disclosures}: None reported.

\paragraph{Glossary}

\begin{center}
\begin{tabularx}{\textwidth}{@{}lX@{}}
\toprule
Key information & The information that can be used to answer the user query\\\midrule
Key document & The unit of text that contains the key information\\\midrule
Spotlight position & Positions where LLMs allocate more attention to than others, typically at the top or bottom of the ranking\\\midrule
Closed-book & Experiment setup where LLMs are provided with user instruction and query only\\\midrule
Chain-of-Thought (CoT) & A prompting technique that instructs LLMs to not only answer the question but also explain the answers step-by-step\\\midrule
Oracle & Experiment setup where LLMs are provided with user instruction, query, and the key information; this is an ideal RAG scenario where the upstream search engine only returns the key information and no extra data.\\\midrule
$Top\_k$ & Number of documents to include in the context of LLM prompt. The search engine returns a list of documents ranked by relevance to the user query. We selected the most related ones.\\\midrule
Conflict information & The content of external documents conflicts with another\\
\bottomrule
\end{tabularx}
\end{center}

\bibliographystyle{medline-url}
\bibliography{ref}

\newpage
\appendix
\setcounter{table}{0}
\setcounter{figure}{0}
\renewcommand\figurename{Supplementary Figure} 
\renewcommand\tablename{Supplementary Table}

\section*{Supplementary materials}

\begin{center}
\captionof{table}{Relationship between QA accuracy and positions of key information in the LLM context. BC - BriefContext. RAG - Retrieval-augmented generation.}
\label{sup tab:Relationship}
\begin{tabular}{lrrrrr}
\toprule
\multirow{4}{*}{Model} & \multirow{4}{*}{\makecell[r]{Position of\\ Key \\Information}} & \multicolumn{4}{c}{\# Documents ($top\_k$)}\\
\cmidrule{3-6}
 & & \multicolumn{2}{c}{8} & \multicolumn{2}{c}{16} \\
\cmidrule(rl){3-4}\cmidrule(rl){5-6}
 & & RAG & BC & RAG & BC \\
\midrule
Mixtral-7x8b & 0th & 59.05 & 57.14 & 61.90 & 62.86 \\
 & 25th & 56.19 & 56.20 & 54.29 & 59.05 \\
 & 50th & 55.24 & 57.14 & 52.38 & 57.14 \\
 & 75th & 56.19 & 58.10 & 56.19 & 60.00 \\
 & 100th & 57.14 & 59.05 & 57.14 & 60.95 \\
\midrule
GPT-3.5-turbo-0125 & 0th & 61.90 & 56.59 & 55.24 & 59.05 \\
 & 25th & 48.57 & 59.05 & 49.52 & 57.14 \\
 & 50th & 46.67 & 60.95 & 48.57 & 59.05 \\
 & 75th & 50.48 & 60.00 & 48.57 & 59.05 \\
 & 100th & 59.05 & 60.95 & 58.10 & 60.00 \\
\bottomrule
\end{tabular}
\end{center}

\newpage

\begin{center}
\captionof{table}{Number of cases with conflict information provided to LLMs and a number of resolved conflicting cases.}
\label{sup tab:Number}
\begin{tabular}{l@{}rrrrrrr}
\toprule
\multirow{3}{*}{Model} & \multirow{3}{*}{\makecell[r]{\# \\Documents \\($top\_k$)}} & \multirow{3}{*}{\makecell[r]{Position of Key\\Information \\(\%)}} & \multicolumn{3}{l}{\makecell[c]{Comparison with Vanilla\\RAG Responses}} & \multirow{3}{*}{\makecell[r]{\#\\Resolved\\Cases}} & \multirow{3}{*}{\makecell[r]{\# Cases with\\Conflicting\\Responses}} \\
\cmidrule{4-6}
 &  &  & Win & Tie & Lose &  &  \\
\midrule
Mixtral-7x8b & 8 & 0th & 13.04 & 78.26 & 8.70 & 17 & 23 \\
 &  & 25th & 30.00 & 60.00 & 10.00 & 15 & 20 \\
 &  & 50th & 28.57 & 61.90 & 9.52 & 16 & 21 \\
 &  & 75th & 38.10 & 47.62 & 14.29 & 16 & 21 \\
 &  & 100th & 22.73 & 72.73 & 4.55 & 18 & 22 \\
 \cmidrule{2-8}
 & 16 & 0th & 8.70 & 86.96 & 4.35 & 20 & 23 \\
 &  & 25th & 42.86 & 47.62 & 9.52 & 16 & 21 \\
 &  & 50th & 40.00 & 40.00 & 20.00 & 19 & 25 \\
 &  & 75th & 42.86 & 47.62 & 9.52 & 17 & 21 \\
 &  & 100th & 45.00 & 45.00 & 10.00 & 17 & 20 \\
\midrule
GPT-3.5-turbo-0125 & 8 & 0th & 5.71 & 74.29 & 20.00 & 22 & 35 \\
 &  & 25th & 36.36 & 48.48 & 15.14 & 23 & 33 \\
 &  & 50th & 28.57 & 67.86 & 3.57 & 18 & 28 \\
 &  & 75th & 33.33 & 54.55 & 12.12 & 23 & 33 \\
 &  & 100th & 23.53 & 58.82 & 17.65 & 25 & 34 \\
 \cmidrule{2-8}
 & 16 & 0th & 15.62 & 84.38 & 0.00 & 26 & 32 \\
 &  & 25th & 28.12 & 62.50 & 9.38 & 24 & 32 \\
 &  & 50th & 28.00 & 60.00 & 12.00 & 15 & 25 \\
 &  & 75th & 42.86 & 42.86 & 14.28 & 23 & 28 \\
 &  & 100th & 30.30 & 54.55 & 15.15 & 26 & 33\\
\bottomrule
\end{tabular}
\end{center}

\newpage

\begin{center}
\captionof{table}{Integration testing of BriefContext with different backbone LLMs. CoT – Chain of thought. RAG - Retrieval-augmented generation. BC – BriefContext.}
\label{sup tab:Integration}
\begin{tabular}{llrrrr}
\toprule
\multirow{2}{*}{Model} & \multirow{2}{*}{Method} & \multicolumn{3}{c}{Dataset} & \multirow{2}{*}{Avg.} \\
\cmidrule{3-5}
 &  & MedMCQA & PubMedQA & BioASQ &  \\
\midrule
LLama2-70B-chat & CoT & 42.60 & 42.20 & 61.17 & 48.66 \\
 & RAG & 43.08 & 50.40 & 73.95 & 55.81 \\
 & BC & 52.00 & 63.60 & 83.82 & 66.47 \\
\midrule
LLama3-70B-instruct & CoT & 70.93 & 59.00 & 83.01 & 70.98 \\
 & RAG & 68.67 & 71.60 & 89.97 & 76.75 \\
 & BC & 68.29 & 79.00 & 89.81 & 79.03 \\
\midrule
Mixtral-7x8b & CoT & 56.28 & 35.20 & 77.51 & 56.33 \\
 & RAG & 56.42 & 67.60 & 87.54 & 70.52 \\
 & BC & 58.60 & 67.40 & 90.61 & 72.20 \\
\midrule
\multirow{3}{*}{GPT-3.5-turbo-0125} & CoT & 55.25 & 36.00 & 74.27 & 55.17 \\
 & RAG & 54.41 & 67.40 & 85.76 & 69.19 \\
 & BC & 54.35 & 75.40 & 87.77 & 72.51\\
 \bottomrule
\end{tabular}
\end{center}

\newpage

\begin{center}
\captionof{table}{Confusion matrix of the “lost-in-the-middle” prediction using IoU of different ranking.}
\label{sup tab:Confusion}
\begin{tabular}{|l|l|l|l|}
\hline
\multicolumn{2}{|c|}{} & \multicolumn{2}{c|}{Truth} \\
\cline{3-4}
\multicolumn{2}{|c|}{} & True & False \\
\hline
\multirow{2}{*}{Prediction} & True & 426 & 34 \\
\cline{2-4}
 & False & 423 & 235\\
\hline
\end{tabular}
\end{center}

\newpage

\begin{center}
\captionof{table}{The characteristics of four datasets used in this study.}
\label{sup tab:dataset}
\begin{tabular}{lrlll}
\toprule
Dataset & \# Questions & Type of Questions & Publicity & Clinical/Biomedical \\
\midrule
MMLU-Med & 1,089 & multiple-choice & Public & Clinical \& Biomedical \\
MedQA-US & 1,273 & multiple-choice & Public & Clinical \& Biomedical \\
MedMCQA & 4,183 & multiple-choice & Public & Clinical \& Biomedical \\
BioASQ-Y/N & 618 & multiple-choice & Public & Biomedical \\
PubMedQA & 500 & multiple-choice & Public & Biomedical \\
MedQ-48 & 48 & open-ended & In-house & Clinical\\
\bottomrule
\end{tabular}
\end{center}

\newpage

\begin{center}
\captionof{table}{Components of the external biomedical knowledge bases.}
\label{sup tab:Components}
\begin{tabular}{lr}
\toprule
PubMed &  \\
\hspace{2em}Number of articles & 23.9M \\
\hspace{2em}Number of snippets & 23.9M \\
\hspace{2em}Tokens, avg & 296 \\
\midrule
MedQA &  \\
\hspace{2em}Number of books & 18 \\
\hspace{2em}Number of snippets & 125.8K \\
\hspace{2em}Tokens, avg & 182\\
\bottomrule
\end{tabular}
\end{center}

\newpage

\begin{center}
\begin{algorithm}
\captionof{algorithm}{ContextMap: divides the context into splits}
\label{alg:map}
\hspace*{\algorithmicindent} \textbf{Input:} context\_data, batch\_size
\begin{algorithmic}[1]

\State requests $\leftarrow$ []
\For {i $\leftarrow$ 0 to length(data) with step batch\_size}
    \State request $\leftarrow$ create\_extraction\_prompt(context\_data[i:i+batch\_size])
    \State requests.append(request)
\EndFor

\State \Return requests
\end{algorithmic}
\end{algorithm}
\end{center}

\newpage

\begin{center}
\begin{algorithm}
\captionof{algorithm}{ContextReduce: aggregates context information}
\label{alg:reduce}
\hspace*{\algorithmicindent} \textbf{Input:} context\_data, batch\_size, model\_id
\begin{algorithmic}[1]

\State requests $\leftarrow$ ContextMap(context\_data, batch\_size)
\State completions $\leftarrow$ []
\For {each request in requests}
    \State completion $\leftarrow$ prompt\_foundation\_model(request, model\_id)
    \State completions.append(completion)
\EndFor

\State summary\_request = create\_summarization\_prompt(completions)
\State final\_answer $\leftarrow$ prompt\_foundation\_model(summary\_request, model\_id)

\State \Return final\_answer
\end{algorithmic}
\end{algorithm}
\end{center}

\end{document}